\documentclass[conference]{IEEEtran}
\IEEEoverridecommandlockouts
\usepackage{cite}
\usepackage{amsmath,amssymb,amsfonts}
\usepackage{multirow}
\usepackage{array}
\usepackage{algorithm2e}
\usepackage{algorithmicx}
\usepackage{graphicx}
\usepackage{float}
\usepackage{silence}
\usepackage{subcaption}
\usepackage{textcomp}
\usepackage{tabularx}
\usepackage{xcolor}
\usepackage{eso-pic}
\usepackage{calc}

\begin{document}

\title{AIOps-Driven Enhancement of Log Anomaly Detection in Unsupervised Scenarios\\}

\author{
\IEEEauthorblockN{Daksh Dave}
\IEEEauthorblockA{\textit{Dept. of Electrical Electronics}\\
\textit{BITS Pilani}\\
Pilani, India\\
f20180391@pilani.bits-pilani.ac.in}\\
\IEEEauthorblockN{Sahil Nawale}
\IEEEauthorblockA{\textit{Information Technology}\\
\textit{Sardar Patel Institute of Technology}\\
Mumbai, India\\
sahil.nawale@spit.ac.in}\\   
\and
\IEEEauthorblockN{Gauransh Sawhney}
\IEEEauthorblockA{\textit{Dept. of Electrical Electronics}\\
\textit{BITS Pilani}\\
Pilani, India\\
f20180325@pilani.bits-pilani.ac.in}\\
\IEEEauthorblockN{Pushkar Aggrawal}
\IEEEauthorblockA{\textit{Dept. of Electrical Electronics}\\
\textit{BITS Pilani}\\
Pilani, India\\
f20180431@pilani.bits-pilani.ac.in}
\and
\IEEEauthorblockN{Dhruv Khut}
\IEEEauthorblockA{\textit{Information Technology}\\
\textit{Sardar Patel Institute of Technology}\\
Mumbai, India\\
dhruv.khut@spit.ac.in}\\   
\IEEEauthorblockN{Prasenjit Bhavathankar}
\IEEEauthorblockA{\textit{Dept. of Computer Engineering}\\
\textit{Sardar Patel Institute of Technology}\\
Mumbai, India\\
p\_bhavathankar@spit.ac.in}\\   
}

\IEEEoverridecommandlockouts
\IEEEpubid{\makebox[\columnwidth]{979-8-3503-1324-6/23/\$31.00~\copyright2023 IEEE\hfill} \hspace{\columnsep}\makebox[\columnwidth]{ }}

\maketitle
\IEEEpubidadjcol

\begin{abstract}
Artificial intelligence operations (AIOps) play a pivotal role in identifying, mitigating, and analyzing anomalous system behaviors and alerts. However, the research landscape in this field remains limited, leaving significant gaps unexplored. This study introduces a novel hybrid framework through an innovative algorithm that incorporates an unsupervised strategy. This strategy integrates Principal Component Analysis (PCA) and Artificial Neural Networks (ANNs) and uses a custom loss function to substantially enhance the effectiveness of log anomaly detection. The proposed approach encompasses the utilization of both simulated and real-world datasets, including logs from SockShop and Hadoop Distributed File System (HDFS). The experimental results are highly promising, demonstrating significant reductions in pseudo-positives. Moreover, this strategy offers notable advantages, such as the ability to process logs in their raw, unprocessed form, and the potential for further enhancements. The successful implementation of this approach showcases a remarkable reduction in anomalous logs, thus unequivocally establishing the efficacy of the proposed methodology. Ultimately, this study makes a substantial contribution to the advancement of log anomaly detection within AIOps platforms, addressing the critical need for effective and efficient log analysis in modern and complex systems.
\end{abstract}


\section{Introduction}
Effective IT (Information technology) operations management relies on detecting and resolving anomalous logs. AI (Artificial Intelligence) for IT Operations (AIOps) leverages log processing and machine learning techniques as promising solutions. ITOps operators utilize log, trace, and telemetry data to improve their offerings  \cite{levin2019aiops}. Anomaly detection in ITOps is critical and can be enhanced through AI algorithms, enabling continuous and automated detection of anomalous logs while mitigating potential hazards \cite{an2022real}. Understanding system failures and their structure is vital in the software development lifecycle \cite{givental2022hybrid}.
Managing cloud-based systems presents challenges, as outages can result in substantial losses. Hence, timely detection of anomalies before service impact becomes crucial. Anomaly detection aims to provide advanced warnings to onsite reliability engineers for prompt diagnosis and intervention \cite{hwang2021fixme}. Automation is crucial in maintaining a continuous log stream for valuable system health information \cite{meng2019loganomaly}. This study focuses exclusively on the unsupervised detection of recurring anomalies. We extract anomalous time windows from collected data and employ a recurring anomaly algorithm for identification. Existing log anomaly detection systems often overlook temporal evolution, focusing only on anomaly occurrence and context. Understanding the origin and temporal patterns of anomalies enhances their significance and impact on system performance. Our study explores strategies that analyze accumulated logs and telemetry data to optimize systems \cite{levin2019aiops}. Additionally, identifying the reasons and structure of system failures is a critical step in the anomaly detection process \cite{givental2022hybrid}.

Machine learning-based methods and AI technologies are essential to apply log anomaly detection effectively in large IT systems. While clustering-based techniques exist, their practical applicability is limited to tabulated data training \cite{givental2022hybrid}. In this study, our algorithm aims to eliminate pseudo-signals and short-term anomalies, improving the signal-to-noise ratio. We will demonstrate the effectiveness of our approach using simulated and real-world datasets, comparing the results with existing anomaly detection systems.

The major contributions of this paper are as follows:
\begin{enumerate}
\item	We introduced an efficient hybrid framework that effectively eliminates pseudo-log anomalies, optimizing operating systems and eliminating the need for data conversion processes, resulting in time and resource savings.
\item   By directly applying the system to anomalous datasets with their original metrics, the approach adapts to different formats and facilitates the identification of logs deviating from average values. This enhances accurate anomaly detection and contributes to improved system stability.
\item	The proposed ensemble approach combines Principal Component Analysis (PCA) and Artificial Neural Networks (ANN) alongwith a custom loss function in an unsupervised strategy for enhanced log analysis. The algorithm achieves significant reductions of in pseudo positives in SockShop and in HDFS, enhancing the accuracy of log anomaly detection. It also highlights the behavior of genuine anomalies, referred to as "pseudo positives" in the context of the research.
\end{enumerate}

\section{Literature Review}
\label{sec2} The emergence of Artificial Intelligence for IT Operations (AIOps) marked a pivotal shift, promising to enhance the accuracy, efficiency, and cost-effectiveness of log anomaly detection.
These advancements in log anomaly detection have been driven by extensive research efforts. An et al. \cite{an2022real} introduced "Log Anomaly Detection techniques," achieving a remarkable 60\% improvement in F1 scores. Meng et al. \cite{meng2019loganomaly} focused on minimizing false signals, while Bogatinovski et al. \cite{bogatinovski2020self,bogatinovski2022leveraging} delved into distributed tracing techniques. 

Further refinements emerged as Guhathakurta et al. \cite{guhathakurta2022utilizing} successfully reduced false log anomalies by 28\%, and Bogatinovski et al. \cite{bogatinovski2022leveraging} proposed root cause analysis. Comprehensive reviews by Jia et al. \cite{jia2021all} and Nedelkoski et al. \cite{nedelkoski2019anomaly} underscored the potential for future advancements.
Additionally, Gholamian \& Ward's study \cite{gholamian2021comprehensive} provided valuable insights into log anomaly detection, while Guo et al. \cite{guo2021detecting} introduced a promising machine learning-based system. Despite these achievements, challenges persist. Some machine learning-based systems can be prohibitively costly \cite{falt2021machine} and may not completely eliminate log anomalies \cite{sauginda2020deep}. AIOps, which leverages log processing, machine learning, and advanced analysis, aims to optimize and transform IT operations \cite{levin2019aiops}.

A variety of techniques have been explored, including the use of vectorized logs and natural language processing methods like word2vec and LogEvent2vec \cite{guo2021translog}.  Additionally, Yang et al. proposed deep-learning techniques for continuous software artifact optimization \cite{yang2021quality}. Recent studies, such as \cite{wittkopp2023pull}, have embraced iterative log analysis through the PULL method to enhance anomaly detection. Furthermore, \cite{cheng2023ai} delves into the broader scope of AI for IT Operations (AIOps) on cloud platforms, shedding light on challenges, opportunities, and emerging trends. \cite{landauer2023deep} demonstrates the capability of AI-based deep learning models, including autoencoders and generative adversarial networks, to detect anomalous logs. In line with this progress, our study makes a notable contribution by introducing a novel system that has demonstrated impressive success in detecting log anomalies.
\begin{figure*}[htbp]
\centerline{\includegraphics[width = 0.6\textwidth]{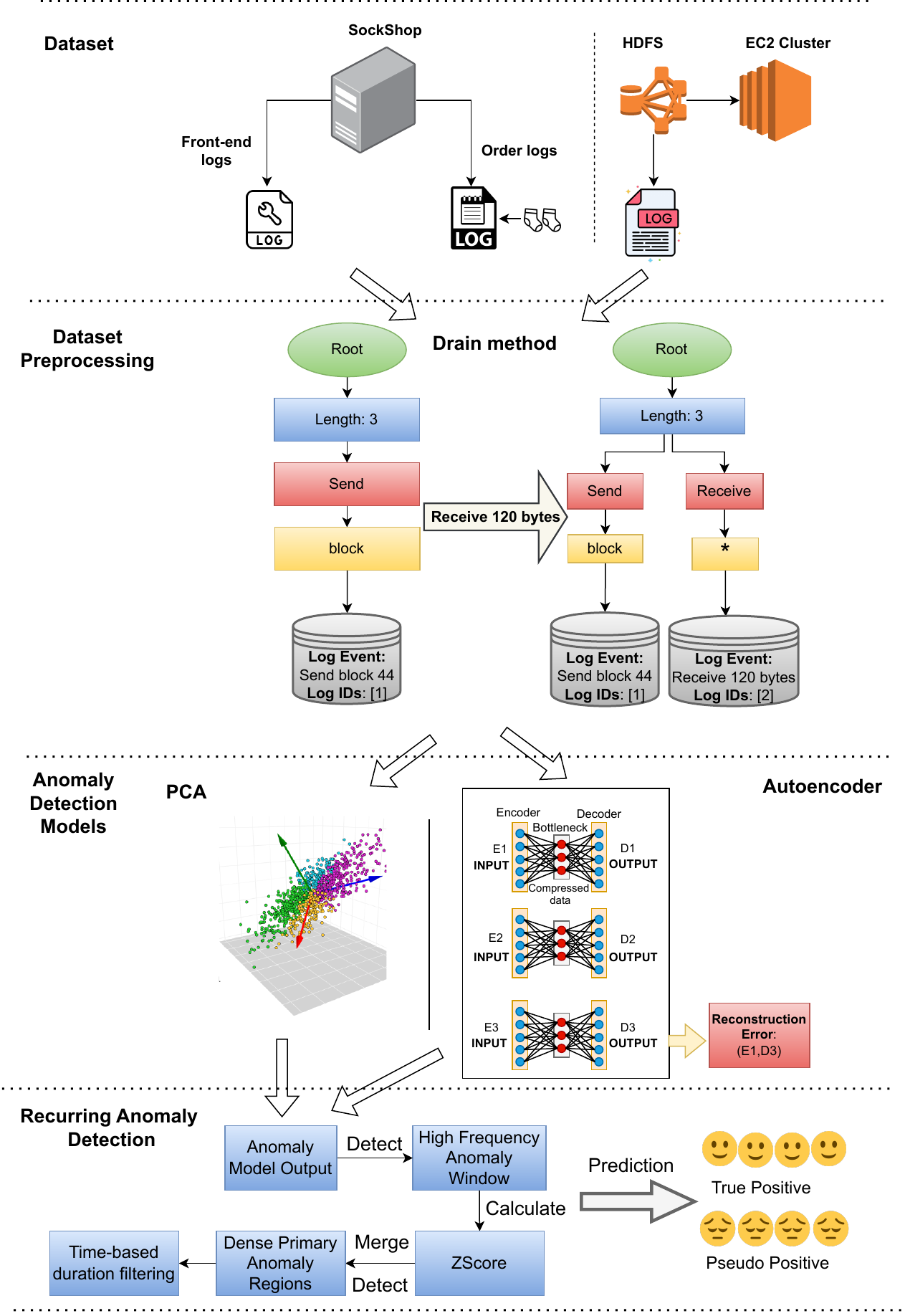}}
\caption{Proposed Framework For Anomaly Detection}
\label{fig:1}
\end{figure*}

\section{Methodology}
We present an innovative methodology that leverages the synergy between Artificial Neural Networks (ANN) and Principal Component Analysis (PCA) to effectively identify and flag log anomalies. To provide a visual overview of our proposed approach, please refer to Figure \ref{fig:1}. Detailed elucidation of our framework can be found in the subsequent sections.
\label{sec3}
\subsection{Data Preprocessing}
In the initial step of data preparation, healthy log data is collected from a period when the cloud-based application was operating optimally. These logs serve as a reference for the anomaly detection process, enabling the identification of deviations from normal system behavior. The data preparation step can be further broken down into the following sub-steps:
\begin{enumerate}
\item Mapping the Logs to a Template: In this stage, the logs are mapped to a template, representing a standardized version of the code appearing in raw logs under different parameters and during various operational times.
\item Creating Time Windows: Each mapped log is then assigned to a fixed 1-second time window. This segmentation allows for more granular log data analysis within specific time intervals.
\item Establishing a Time Vector Template: The mapped logs within each time window are aggregated to create a template for the time vector count to facilitate further analysis. This template captures the overall patterns and characteristics of the log data during the designated time interval.
\end{enumerate}

The time vector is denoted by (\ref{eq}) as 
\begin{equation}
\label{eq}
T=\left[t_1, t_2, t_3, \ldots, t_n\right]
\end{equation}
Here, $T$ denotes the time vector template, and $t_1, t_2, t_3, \ldots, t_n$ represents the individual time values within the vector. The time vector template captures a sequence of time values or timestamps for analysis.

\subsection{Anomaly Detection Models}
\subsubsection{PCA System Employment}
The methodology's next step is utilizing the Principal Component Analysis (PCA) system. This algorithm plays a crucial role in the assessment process by providing enhanced analysis capabilities through dimensionality reduction and anomaly detection, as highlighted by \cite{dang2019aiops} in their research. Unlike many other systems, PCA offers the ability to identify patterns and establish a baseline for what constitutes a "normal class" within the log data.

The PCA system can effectively detect anomalous behavior by employing distance metrics and analyzing the patterns formed. This allows for the identification of deviations from the established normal class, aiding in the early detection of potential issues or anomalies within the system.

Once the data preparation steps are completed, the PCA system is trained using the previously prepared log data denoted by the symbol $L$. This training process enables the PCA system to learn and identify patterns, making it capable of accurately detecting anomalies in subsequent log data. The first step involves calculating the Covariance matrix through \eqref{eq3}. The covariance matrix ($C$) is calculated by multiplying the transpose of the log data matrix ($L'$) with the original data matrix ($L$), divided by ($n-1$), where ($n$) represents the number of observations.
\begin{equation}
\label{eq3}
C=1 /(n-1) * L' * L
\end{equation}

After obtaining the covariance matrix, the next step is to find the eigenvector matrix through \eqref{eq4}. The eigenvalue-eigenvector equation shows that the covariance matrix ($C$) multiplied by the eigenvector matrix ($V$) is equal to the eigenvectors ($V$) scaled by the corresponding eigenvalues ($\lambda$).
\begin{equation}
\label{eq4}
C * V=\lambda * V
\end{equation}

This is followed by decomposing the original log data matrix ($L$) into the product of the left singular vector matrix ($U$), the diagonal matrix of singular values ($\sum$), and the transpose of the right singular vector matrix ($V'$) as denoted by \eqref{eq5}:
\begin{equation}
\label{eq5}
L=U * \Sigma * V'
\end{equation}

We then proceed with projecting the data onto the Principal Components to form the transformed data matrix ($Y$), representing the data in terms of the principal components. The matrix is obtained by projecting the original data matrix ($L$) onto the matrix of eigenvectors ($V$), as given by \eqref{eq6}:
\begin{equation}\label{eq6}
Y=L* V
\end{equation}

\subsubsection{Employing the Auto encoder-based Anomaly Detector}
In the methodology, the next step involves implementing the Autoencoder-based Anomaly Detector, a neural network designed to operate unsupervised. The primary objective of the autoencoder is to extract the principal patterns from the normal data by reducing its dimensionality. By leveraging this capability, dimensional measurements can effectively detect anomalies.

To achieve this, the autoencoder network consists of three encoders and three decoders, working in tandem to capture the essential characteristics of the input data and reconstruct it accurately. The encoder function for the $i^{\mathrm{th}}$ encoder is represented by \eqref{eq7}
\begin{equation}\label{eq7}
E_i=f\left(W_i * x+b_{i}\right)
\end{equation}
where $E_i$ is the output of the $i^{\mathrm{th}}$ encoder, $x$ is the input data, $W_{i}$ is the weight matrix, $b_{i}$ is the bias term and $f(\cdot)$ denotes the activation function. Similarly, the decoder function for the $i^{\mathrm{th}}$ decoder is represented by \eqref{eq8}
\begin{equation}
\label{eq8}
D_{i}=f(W_{i} * E_{i}+b_{i})
\end{equation}
 where $D_{i}$ is the output of the $i^{\mathrm{th}}$ decoder, $E_{i}$ is the output of the corresponding encoder, $W_{i}$ is the weight matrix, $b_{i}$ is the bias term, and $f(\cdot)$ represents the activation function.

The reconstruction error measures the discrepancy between the input data and its reconstructed output. The reconstruction error, denoted by $R$, is typically calculated using a loss function such as mean squared error (MSE). For a given input $x$, the reconstruction error is given by \eqref{eq9}
\begin{equation}
\label{eq9}
R=\mathrm{MSE}\left(x, D_3\right)
\end{equation}
 where $D_3$ represents the output of the third decoder. $\mathrm{MSE}\left(x, D_3\right)$ quantifies the mean squared error between the input $x$ and its reconstructed output $D_3$. By analyzing the reconstruction errors, higher values of $R$ indicate the presence and magnitude of anomalies within the dataset, enabling effective anomaly detection using the Autoencoder-based Anomaly Detector.
\subsubsection{Custom Loss Function}
Central to our approach is the design of a custom loss function, denoted as $L_{\text{custom}}$, tailored explicitly for log anomaly detection. The custom loss function is formulated as follows:
\begin{equation}
L_{\text{custom}} = \sum \left(\alpha \cdot |x - \hat{x}| + \beta \cdot |PCA(x) - PCA(\hat{x})|\right)
\end{equation}
Here, $x$ represents the original log entry, $\hat{x}$ denotes the reconstructed log entry from the autoencoder, and $PCA(x)$ and $PCA(\hat{x})$ represent the principal component scores of the original and reconstructed log entries, respectively. The parameters $\alpha$ and $\beta$ are tuned to control the relative importance of the reconstruction error and PCA-based anomaly scores.

\subsubsection{Ensemble Framework}
In the ensemble framework, we combine the outputs of PCA-based anomaly detection and the autoencoder-based approach using a weighted sum:
\begin{equation}
\text{Ensemble Score} = w_{\text{PCA}} \cdot PCA(x) + w_{\text{Autoencoder}} \cdot L_{\text{custom}}
\end{equation}
The ensemble score integrates PCA's statistical properties and the enhanced feature learning from the autoencoder. Weight parameters $w_{\text{PCA}}$ and $w_{\text{Autoencoder}}$ are optimized during training to achieve the best trade-off between the two methods.

\subsection{Recurring Anomaly Detection}
The subsequent critical step in our model is the detection of recurring anomalies. Recurring anomalies are characterized as anomalies that exhibit regular frequency and persist within a system for extended periods. To identify these recurring anomalies, we utilize the output generated by a generic anomaly detector, which helps identify the time periods when these anomalies occur.

The identified time periods serve as windows for the recurring anomaly detection algorithm. By employing a sliding window approach, we abstract the results and assess the occurrence frequency of the anomalous log labels within these windows.

The first step in this process involves identifying all the regions with high-frequency anomalies. We then shift our focus to the denser areas, where the recurrence frequency is highest, and use it as input for further analysis. In the third step, we merge these dense anomalous regions and perform a time duration check to ensure that the anomalies persist over a significant period. These steps are elaborated below:
\begin{table*}[!t]
\caption{Anomaly Detection Results for SockShop and HDFS Data using PCA and Auto Encoder Detectors}
\centering
\begin{tabular}{|l|c|c|c|c|c|c|c|}
\hline
\multirow{2}{*}{\textbf{METHODS}} & \multicolumn{2}{|c|}{\begin{tabular}[c]{@{}c@{}}\textbf{SOCK SHOP}\\ {[}Front End{]}\end{tabular}} & \multicolumn{2}{|c|}{\begin{tabular}[c]{@{}c@{}}\textbf{SOCK SHOP}\\    {[}Orders{]}\end{tabular}} & \multicolumn{2}{|c|}{\textbf{HDFS}} \\ 
\cline{2-7} 
 & \multicolumn{1}{|c|}{\textbf{\textit{PPR }}} & \textbf{\textit{TPR}} & 
 \multicolumn{1}{|c|}{\textbf{\textit{PPR}}} & \textbf{\textit{TPR}}
 & \multicolumn{1}{|c|}{\textbf{\textit{PPR}}} & \textbf{\textit{TPR}} \\ 
 \hline
{[}PCA{]}-   High-Frequency Window Detection & \multicolumn{1}{|c|}{0} & 0 & \multicolumn{1}{|c|}{77.777\%} & 0 & \multicolumn{1}{|c|}{16.87\%} & 0.75\% \\ \hline
{[}PCA{]} –   Dense Anomalous Region Detection & \multicolumn{1}{|c|}{1.26\%} & 0\% & \multicolumn{1}{|c|}{77.77\%} & 0 & \multicolumn{1}{|c|}{16.87\%} & 0.75\% \\ \hline
{[}PCA{]}   Duration-Based filtering & \multicolumn{1}{|c|}{27.84\%} & 3.25\% & \multicolumn{1}{|c|}{77.77\%} & 7.84\% &  \multicolumn{1}{|c|}{28.75\%} & 0.38\% \\
\hline
{[}Auto   Encoder{]} – High-Frequency Window Detection & \multicolumn{1}{|c|}{2.85\%} & 9.37\% & \multicolumn{1}{|c|}{9.09\%} & 0 &  \multicolumn{1}{|c|}{3.28\%} & 0.01\% \\
\hline
{[}Auto   Encoder{]} – Dense Anomalous Region Detection & \multicolumn{1}{|c|}{2.85\%} & 10.93 \%  & \multicolumn{1}{|c|}{9.09\%} & 0 & \multicolumn{1}{|c|}{5.03\%} & 0\\ \hline
{[}Auto   Encoder{]} – Duration-Based Filtering & \multicolumn{1}{|c|}{20.0\%} & 23.43\% &  \multicolumn{1}{|c|}{45.45\%} & 0 & \multicolumn{1}{|c|}{9.40\%} & 0 \\ 
\hline
{[}Ensemble{]} – High-Frequency Window Detection & \multicolumn{1}{|c|}{\textbf{4.65\%}} & 0.94\% & \multicolumn{1}{|c|}{\textbf{81.14\%}} & 0.23\% &  \multicolumn{1}{|c|}{\textbf{23.92\%}} & 0.49\% \\
\hline
{[}Ensemble{]} – Dense Anomalous Region Detection & \multicolumn{1}{|c|}{\textbf{4.23\%}} & 0.56 \%  & \multicolumn{1}{|c|}{\textbf{79.37\%}} & 0 & \multicolumn{1}{|c|}{12.46\%} & 0.11\\ \hline
{[}Ensemble{]} – Duration-Based Filtering & \multicolumn{1}{|c|}{\textbf{30.07\%}} & \textbf{1.95\%} &  \multicolumn{1}{|c|}{\textbf{89.32\%}} & 0 & \multicolumn{1}{|c|}{\textbf{29.35\%}} & 0.89\% \\ 
\hline
\end{tabular}
\label{tab1}
\end{table*}

\subsubsection{Detection of High-frequency Anomalous Windows}
In order to improve the accuracy of anomaly detection, the next step in the methodology is to identify high-frequency anomalous windows, which exhibit a significant deviation from the mean occurrence time during normal system operation. These windows represent optimal zones for detection and require further investigation.


A robust peak detection strategy is employed to reduce the number of windows that need to be analyzed and enhance accuracy. We use an algorithm to signal a peak, when a new data point deviates by a threshold number of standard deviations from the moving mean. In this study, we have adopted Brakel's algorithm \cite{van2014robust} due to its flexibility and parameter manipulations, which enhance its performance. The algorithm includes the Lag, Threshold and Influence which can be utilized for peak detection.



\subsubsection{Primary Detection of Dense Anomalous Regions}

To identify the primary dense anomalies, a moving window approach is used to analyze the high-frequency anomalous region. Within this region, a window of size N, denoted as d, is employed to detect the dense anomalies. This step aims to differentiate and exclude short-term or transient anomalies that have minimal impact on the system.

The primary dense anomalies refer to the anomalies that exhibit a higher concentration and persistence within the analyzed data. Denser patterns characterize these anomalies and are more likely to indicate noteworthy system abnormalities. By focusing on these primary dense anomalies, we can prioritize detecting and analyzing anomalies that have a greater impact on the system's performance and reliability.

\subsubsection{Time-Based Elimination}
The “Time-Based Elimination" step consists of two integral parts. In the first part, we focus on identifying dense regions encompassed or surrounded by other dense areas. These identified dense regions are then consolidated to form a more comprehensive dense region. To illustrate this process, let's consider a scenario where a dense region A is close to another dense region B. If the distance separating them is within a specified threshold value of N, they can be merged together to create one consolidated region. The condition merging the two dense regions can be denoted by \eqref{eq11},
\begin{equation}
\label{eq11}
\left(W_a-W_b\right)<N
\end{equation}

Where $W_a$ is the dense region $A$ and $W_b$ is the dense region $B$ and $N$ is the specified threshold distance separating the two regions.

Moving on to the second part of this step, we examine the temporal aspect by evaluating the duration of the consolidated region. We ensure that the duration does not exceed a predefined length $l$, representing the minimum duration required for an anomalous event to be considered significant. This duration can be determined based on the time it takes for the anomaly in the cloud application to propagate and eventually reach the end user. Considering the spatial relationship and the temporal characteristics, this step plays a vital role in refining and validating the detected dense anomalies. It ensures that the identified anomalies are spatially dense and exhibit a temporal coherence that aligns with the expected behavior of anomalous events. This further enhances the accuracy and relevance of the detected anomalies within the given context.

\section{Experimental Setup and Results}
\label{sec4}
In the preliminary setup of our experimentation, we utilized a simulated dataset obtained from the "Sock Shop," an online application specializing in selling socks. This application is built on OpenShift and leverages an Istio mesh for enhanced functionality \cite{van2014robust}. The Sock Shop comprises seven microservices that communicate with each other using the "REST over HTTPS" format.

The Istio mesh services were configured to generate 2xx, 4xx an 5xx error codes to simulate real-world system anomalies. This deliberate introduction of anomalies allowed us to create a test environment that closely resembled actual system behavior. The primary dataset used in our experiments consisted of logs collected over 24 hours. These logs captured both normal and anomalous time behavior and served as the foundation for training the anomaly detectors over a week.

Furthermore, we incorporated a real-world HDFS log dataset from Hadoop clusters running on 612 EC2 nodes. This dataset contained a staggering 89,478,103 million log lines collected over a period of 7 days. The inclusion of this dataset in our setup added a valuable real-world dimension to our experiments. All log segments that represented real-time run errors were manually labeled to establish ground truth for evaluation purposes to distinguish them as anomalies.

In the secondary experiment setup, we utilized the 'LogParse' method \cite{log-parser} for raw log extraction. This method employs templatization to parse and categorize logs, with a 1-second time window for precise segmentation and synthesis. To perform anomaly detection, we employed a custom Autoencoder model based on the TensorFlow Library, providing a robust alternative to commercial solutions.

The effectiveness of the anomaly detection process relies on two key parameters: the density threshold and the lag. Adjusting these parameters can be crucial in achieving optimal results. A higher lag value can be chosen if we anticipate faults occurring after the experiment, as it helps establish a normal behavior pattern. Conversely, the lag value can be reduced if the erroneous log is introduced simultaneously with the collected data.
A density threshold of 45\% was found to work well for both datasets. Key parameters, such as a window size (W) set at 2 minutes, a slide size (S) at 45 seconds, influence at 0.5, and z-score ranging from 2.0 to 5.0, were chosen to ensure precise evaluation. To evaluate the effectiveness of the approach, the reduction of pseudo positives (PPR) was compared to the initial number of pseudo alerts. This metric proved to be significant in extracting inappropriate logs. 


Table \ref{tab1} when compared with \cite{AIOps_Copy} presents insights into the reduction of pseudo positives and transient anomalies. When applying the recurring anomaly detection system along with our custom loss function and ensemble auto-encoder to the PCA-based anomaly detector in the SockShop dataset, significant reductions in pseudo positives were achieved.

The bold values in Table-\ref{tab1} represent an improved performance by our framework. Our results section highlights the superiority of our ensemble method in log anomaly detection. Specifically, it achieved a significant 30.07\% reduction in pseudo positives for front-end services and an even more substantial 89.32\% reduction for order services. In the Hadoop Distributed File System (HDFS) dataset, the ensemble consistently reduced pseudo positives by 29.35\%. Additionally, combining the Autoencoder with our ensemble led to notable reductions of 20\%, 45\%, and 9.4\% in pseudo positives for the front-end service and HDFS datasets. The ensemble-based duration filtering method consistently outperformed all other approaches on both SockShop and HDFS datasets, thanks to its high pseudo-positive reduction rate and low transient positive reduction rate. Importantly, these reductions in false positives did not significantly impact True Positive Reduction (TPR) across all methods, affirming the effectiveness of our ensemble approach in enhancing log anomaly detection, a crucial aspect of system security and stability.

One notable feature of the algorithm is its ability to handle different raw system metrics across multiple services without the need for applying an anomaly detector. By introducing a time series of raw data into the pipeline, the algorithm identifies time windows that deviate from the normal mean values. This process allows for the detection of anomalies based on their significant deviations in frequency over an extended time frame. Applying the recurring technique proves to be significant in identifying anomalies that deviate considerably from the mean and exhibit a high incidence of occurrence. By incorporating log-based recurring anomaly detectors, the algorithm effectively reduces the number of pseudo-positives, further improving the accuracy of anomaly detection.

\section{Conclusions}
\label{sec5}
This research elucidated a novel strategy, proficient in markedly reducing pseudo-positives within log datasets, enhancing anomaly detection and system log management in applied instances within the SockShop and HDFS datasets. The substantial merits of the proposed method lie in its significant efficacy in diminishing pseudo-positives, efficient handling of raw anomalous datasets without additional data conversion, and its unsupervised adaptability across various domains. Despite the demonstrated strengths, the model isn’t devoid of challenges; nuanced tuning for optimal performance across varied datasets, scalability apprehensions, and questions about real-time applicability in dynamically rich data environments remain. While providing a robust foundation, the system necessitates ongoing refinement and examination in its evolution, ensuring its sustained applicability and efficiency in forthcoming anomaly detection endeavors, thereby upholding its relevance amidst the swift advancements in data management and analysis in research and practical applications.

\bibliographystyle{ieeetr}
\bibliography{ref}

\end{document}